\documentclass{Interspeech}

\interspeechcameraready

\title{\textsc{HArnESS}: Lightweight Distilled Arabic Speech Foundation Models}

\author{Vrunda N.}{Sukhadia}
\author {Shammur Absar Chowdhury}

\affiliation{Qatar Computing Research Institute}{HBKU} {Qatar}

\email{sukhadiavrunda@gmail.com, shchowdhury@hbku.edu.qa}

\usepackage{color}
\usepackage{float}
\usepackage{tabularray}
\usepackage[table]{xcolor}
\usepackage{booktabs,tabularx}
\usepackage{multirow}

\usepackage{comment}
\usepackage{graphicx}   

\begin{document}

\maketitle

\begin{abstract}

Large pre-trained speech models excel in downstream tasks but their deployment is impractical for resource-limited environments. In this paper, we introduce HArnESS, the first Arabic-centric self-supervised speech model family, designed to capture Arabic speech nuances. Using iterative self-distillation, we train large bilingual HArnESS (HL) SSL models and then distill knowledge into compressed student models (HS, HST), preserving Arabic-specific representations. We use low-rank approximation to further compact the teacher’s discrete supervision into shallow, thin models. We evaluate HArnESS on Arabic ASR, Speaker Emotion Recognition (SER), and Dialect Identification (DID), demonstrating effectiveness against HuBERT and XLS-R. With minimal fine-tuning, HArnESS achieves SOTA or comparable performance, making it a lightweight yet powerful alternative for real-world use. We release our distilled models and findings to support responsible research and deployment in low-resource settings.
\end{abstract}

\section{Introduction}



Self-supervised learning (SSL) revolutionizes speech processing by extracting and learning transferable representations from large amounts of unlabeled speech.
These large SSL models capture rich speech representations, making them highly effective across a wide range of speech-related tasks \cite{chen2022wavlm, hsu2021hubert,baevski2022data2vec,mohamed2022self,chung2021w2v,yang2021superbspeechprocessinguniversal}. These models can be leveraged either as feature extractors or fine-tuned with a small amount of labeled data, significantly improving performance in low-resource settings.

However, the generalization capability of SSL models is highly dependent on the diversity and volume of training data. Multilingual SSL speech models such as XLS-R \cite{babu2021xlsr} have demonstrated superior performance in low-resource languages compared to monolingual models, particularly those trained in high-resource languages like English \cite{shi2023mlsuperbmultilingualspeechuniversal}. Yet, recent findings by \cite{storey2024language} indicate that models like XLS-R tend to prioritize languages with abundant training data, potentially leading to suboptimal performance for underrepresented languages. This suggests that relying solely on multilingual models may not ensure optimal performance across all languages, and language-specific SSL models could be essential to address this gap effectively.

The \textbf{Arabic} language poses a unique challenge in speech processing due to its vast linguistic diversity. Spoken across 22 countries, Arabic includes more than 20 mutually incomprehensible dialects, with significant influences from other languages such as English and French \cite{ali_connecting_2021}. 
Given this linguistic complexity, choosing an appropriate SSL model for Arabic speech tasks requires a model that can effectively capture the nuances of spoken dialects while preserving cultural and phonetic diversity. Multilingual models, though useful, may not fully grasp the intricacies of Arabic speech, making the case for a dedicated Arabic-English SSL model.
However, training and deploying language-specific speech SSL models requires significant computational resources, large-scale unlabeled data, and prolonged training times, making them costly and inaccessible for many researchers. 
These high resource demands hinder efficient fine-tuning and deployment in resource-constrained settings.


\textbf{Knowledge distillation} has emerged as a key technique for compressing large models while preserving their performance. Distillation enables a smaller, more efficient model (student) to learn from a larger, computationally expensive teacher model, reducing memory usage and inference time without significant performance degradation.
Early works such as DistillHuBERT \cite{chang2022distilhubert}, FitHuBERT \cite{lee2022fithubert}, DPHuBERT \cite{peng2023dphubert}, SKILL \cite{zampierin2024skill} among others \cite{ashihara2022deep,wang2022lighthubert} have applied task-agnostic knowledge distillation to HuBERT \cite{hsu2021hubert}.

In this study, we introduce \textbf{H}uBERT-based \textbf{Ar}abic a\textbf{n}d \textbf{E}nglish \textbf{S}elf-\textbf{S}upervised Speech (HArnESS) models, the first family of Arabic self-supervised speech models, jointly trained on large-scale Arabic and English speech data. By training specifically on Arabic and English, HArnESS ensures better adaptation to dialectal variations, and phonetic intricacies, making it a suitable choice for Arabic speech applications. 

The HArnESS models are trained using iterative self-distillation, following the HuBERT architecture. 
This approach enhances model performance by using its own predecessor's predictions as supervisory signals across multiple training iterations. 
We leverage this learning paradigm, and train HArnESS-L with 24 encoder layers in the first few iterations, preserving its deep architecture. In the following iteration, we distill its knowledge into lightweight student models, resulting in HArnESS-S (shallow) and HArnESS-ST (shallow and thin) architectures. 
Furthermore, we apply low-rank approximation to compact the supervisory signal with rich and informative representation and ensure efficient knowledge transfer.



We evaluated the HArnESS-L, HArnESS-S and HArnESS-ST models on downstream tasks including Automatic Speech Recognition (ASR), Speaker emotion recognition (SER), and Dialect identification (DID). We also compared the performance of these models with HuBERT-large (English)  and XLS-R (multilingual) \cite{babu2021xlsr} models.

\begin{figure}[!h]
    \centering    
    \includegraphics[scale=0.55]{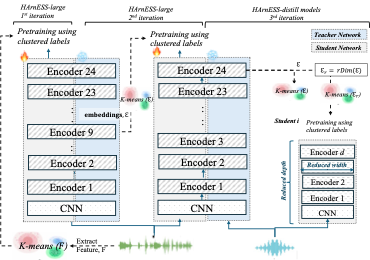}        \vspace{-0.2cm}
    \caption{Overview: Iterative self-distillation training to build HArnESS model family.}
    \label{fig:overview}
    \vspace{-0.2cm}
\end{figure}

\noindent Therefore, our key contributions are:
\begin{enumerate}
    \item Introducing HArnESS, the first Arabic-centric self-supervised speech model family with HArnESS-L (large), HArnESS-S (shallow), and HArnESS-ST (shallow and thin) architectures.
    \item Explore the iterative self-distillation paradigm for Arabic model compression and knowledge distillation.
    \item Investigate the effect of compact supervision signals on model performance. 
    \item Publicly release distilled models (HArnESS-S and HArnESS-ST) for research.\footnote{Links removed for anonymity.}
    \item Benchmark HArnESS on content (ASR), speaker information (DID), and paralinguistic (SER) tasks. 
    \item Release benchmark datasets to support further research.\footnotemark[1]
    
\end{enumerate}

\section{HArnESS Models}
Figure \ref{fig:overview} gives an overview of the HArnESS training pipeline, which follows a HuBERT-style iterative self-distillation approach.

\noindent\textbf{Training Regimen} At each iteration $i$, we train a model $\mathbf{M}_{i}$ using the pseudo-labels from the previous $(i-1)$ iteration model as supervisory signals. 
The model then learns to classify randomly masked frames to these pseudo-labels. 
In the first two iterations, we retain the same architecture to enhance feature abstraction in the student model. From the third iteration, we compress the model through distillation by: \textit{(a)} reducing depth ($d$) for a shallower architecture; \textit{(b)} reducing width ($enc_d$) for a thinner model and
\textit{(c)} lowering attention capacity ($attn$) by decreasing attention heads.

\noindent\textbf{Model Architecture}
The HArnESS model consists of a convolutional (CNN) feature extractor and Transformer encoders. Similar to HuBERT, WavLM \cite{chen2022wavlm} and Wav2Vec2.0 \cite{baevski2020wav2vec}, the CNN feature extractor includes 7 temporal convolutions layers. The transformer encoder layers $l$, has embedding dimension $enc_d$, and are composed of a multi-head self-attention (MHA) with $attn$ heads and a position-wise feed-forward network (FFN).

\noindent\textbf{Training Objectives} We use the weighted sum of two cross-entropy losses, applied separately to the masked and unmasked frames as training objectives.

\noindent\textbf{Weight Initialization} We explored different weight initialization strategies to improve convergence and stability. We experimented with random weight initialization, where model weights were initialized using a uniform random distribution to introduce diversity in parameter values at the start of training. Additionally, we also employed a blocked averaging initialization approach, where weights were initialized by averaging blocks of layers from the preceding model. 

\noindent\textbf{Pseudo-labels generation} 
We employ K-means clustering to assign discrete labels to speech frames, leveraging last-layer embeddings, which capture the most refined high-level representations from the model.\footnote{We also explored averaged embeddings from selected layers but found no significant differences.} To enhance clustering efficiency for distillation, we apply Principal Component Analysis (PCA) to filter out redundant information and retain only the most meaningful features for representation.
For the initial iteration, $i=0$, we initialize training by extracting MFCC features from the raw speech input $x$ and use them to generate the initial pseudo-labels. 





\section{Experimental Setups}

\subsection{Pre-training Data}

\noindent\textbf{Iteration 1 \& 2 Data:} We utilized publicly available Arabic and English speech corpora including QASR \cite{mubarak_qasr_2021}, MGB3 \cite{ali2017speech}, LibriSpeech \cite{panayotov2015librispeech}, Common Voice (English and Arabic) \cite{ardila2020commonvoicemassivelymultilingualspeech}, GigaSpeech \cite{chen2021gigaspeechevolvingmultidomainasr} among others.
To make the model culturally sensitive we incorporate spoken content from 15 Arabic-speaking countries (crawled from YouTube data), covering diverse dialects.
Finally, we augment the selected datasets with approaches including speed perturbation, SpecAugment, adding reverberation and noise among others to regularize and enhance robustness of the model. 
We then pretrain the model on these \textbf{23K hours} of speech, ensuring an almost balanced distribution between Arabic and English language. We strictly exclude any official test and development sets of the datasets, from pretraining to prevent data leakage and to ensure reliable evaluation. For training the k-means model, we used 300 hours subset of the 23K hours data.

\noindent\textbf{Iteration 3 Data:} Our primary goal is to develop a lightweight Arabic-centric model. In this phase, we selected $\approx $\textbf{1,100 hours} of Arabic data from the QASR training dataset for knowledge distillation. To train the k-means model for generating pseudo-labels, we randomly sampled 30\% of the Iteration3 data ($\approx$300 hours), ensuring efficient clustering while maintaining some linguistic diversity.

\begin{table}[h]
\centering
\caption{Downstream Tasks and Dataset Statistics. KE: KSU Emotion dataset.}
\label{tab:down_data_stats}
\scalebox{0.8}{
\begin{tabular}{lcccc}
\toprule
\textbf{Data} & \textbf{Trn(hrs)} & \textbf{Dev(hrs)} & \textbf{Tst(hrs)} & \begin{tabular}{@{}c@{}}\textbf{Task} \\ (\# labels) \end{tabular} \\ \midrule
\textit{QASR}        & 300  & 6.0   & --    & \textbf{ASR} \\
\textit{MGB2}       & --   & --   & 9.57  & -- \\
\textit{MGB3}       & --   & --   & 5.78  & -- \\\midrule 
\textit{KE}  & 3.30   & 0.83  & 1.0    & \begin{tabular}{@{}c@{}} \textbf{SER} (6): \\ Happy, Sad, \\ Surprise, Neutral, \\ Angry, Interrogative \end{tabular} \\ \midrule
\textit{ADI5}       &42.90    & 10.0  & 10.0    &\begin{tabular}{@{}c@{}} \textbf{DID} (5): \\ MSA, EGY, \\ LEV, NOR, GLF \end{tabular} \\
\bottomrule
\end{tabular}
}
\end{table}

\subsection{Downstream Tasks and Data}

Benchmarking speech SSL models for English has been extensively studied, with SUPERB \cite{yang2021superbspeechprocessinguniversal} serving as a key standard for evaluating SSL efficacy across tasks like content recognition, speaker information, and paralinguistics. However, no such standardized benchmark exists for Arabic speech.
To address this gap, we introduce a benchmarking effort for Arabic SSL models, evaluating performance across key tasks: ASR for content recognition, dialect identification (DID) for speaker information, and speaker emotion recognition (SER) for paralinguistic analysis. For ASR, we fine-tune Harness models on a small subset of QASR data (300 hrs) and test on MGB2. We investigate the generalization capability of these small models on unseen data by evaluating the model with MGB3 testset. For SER, we use KSUEmotion \cite{9393909}, a dataset collected from 23 speakers with 6 emotion classes. The dataset is split into train (3.30 h), dev (50 min) and test (1 h).\footnote{We will release the data split for reproducibility.} For DID, we use the ADI5 dataset covering 5 region-based dialect classes. Details in Table \ref{tab:down_data_stats}.
For downstream task evaluation, we opt for widely used measures -- word error rate (WER) for ASR and accuracy (Acc) for DID and SER.

\begin{table*}[!h]
\centering
\caption{Reported Performance Comparison of Distilled Models on ASR, SER, and DID Tasks. L: Large, S: Shallow, ST: S+Thin. $\Delta S$: Overall structural compression. SOTA* models are large models designed for specific tasks and trained with large/full corresponding training data (upper bound performance). } 
\label{tab.rslt}
\vspace{-0.3cm}
\scalebox{0.8}{
\begin{tabular}{l|cc|c|c}
\toprule
\multirow{2}{*}{\textbf{Models}} & \multicolumn{2}{c|} { \begin{tabular}{@{}c@{}} \textbf{ASR (WER ↓)} \\ \textit{(Train: 300 hrs QASR subset)} \end{tabular}} & \textbf{SER (Acc ↑)} & \textbf{DID (Acc ↑)} \\  
\cmidrule(lr){2-3} \cmidrule(lr){4-4} \cmidrule(lr){5-5}
 & \textbf{MGB2} &  \textbf{MGB3} & \textbf{KSUEmotion} & \textbf{ADI5} \\  
\midrule
\textbf{SOTA*}  & 10.24  &  21.31  & 83.31\%  & 82.5\%  \\  \midrule\midrule
\multicolumn{5}{c} {\textit{SSL Large models}}
\\
\midrule
\textbf{HuBERT-L} (\textit{English})  & 22.6  & 51.2  & 91.92\%  & 64.14\%  \\  
\textbf{XLS-R} (\textit{Multilingual}) & 22.60    & 51.80  & 73.32\%  & 42.35\%  \\  
\midrule

\textbf{HArnESS-L} (\textit{Bilingual: Arabic-English})  & \textbf{15.50}   & \textbf{41.60}  & \textbf{94.66\%}  & \textbf{84.98\% } \\  \midrule\midrule
\multicolumn{5}{c} {\textit{Compressed using }$\approx$ \textit{1000h} \textit{Arabic only data}}
\\

\midrule
\textbf{HArnESS-S} ($\Delta S=79.4\%$)  & \textcolor{blue}{\textit{20.20}}   & \textcolor{blue}{\textit{52.80}}  & \textcolor{blue}{\textit{91.15\%}}  & \textcolor{blue}{\textit{70.84\%}}  \\  
\textbf{HArnESS-ST} ($\Delta S=93.7\%$)  &   23.20 & 58.20  & 89.02\%  & 69.77\%  \\  
\textbf{HArnESS-ST$^\Xi$} ($\Delta S=93.7\%$) & 22.50   & 55.60  & 87.34\%  & 61.64\%   \\  
\bottomrule
\end{tabular}}
\vspace{-0.2cm}
\end{table*}

\subsection{Pre-training Training Parameters}
\begin{table}[H]
\centering
\caption{SSL Model Architecture Comparison. XR: XLS-R, HuL: HuBERT-Large, H-L: HArnESS-Large, H-S: HArnESS-Shallow, H-ST: HArnESS-Shallow and Thin. Dim. dimension, Emb.: Embedding. $L*_{emb}$: Embedding from layer * (e.g. 23) of model from iteration $i$. }
\label{tab:arch_desc}
\scalebox{0.7}{
\begin{tabular}{l|cccccc}
\toprule
\textit{Models} & \textbf{XR} & \textbf{HuL} & \textbf{H-L} & \textbf{H-S} & \textbf{H-ST} & \textbf{H-ST (PCA)} \\  
\midrule
\textit{Supervision} 
& -- & -- 
& \shortstack{$L9_{\text{emb}}$\\($i=1$)} 
& \multicolumn{2}{c}{\shortstack{$L23_{\text{emb}}$\\($i=2$)}} 
& \shortstack{PCA($L23_{\text{emb}}$)\\($i=2$)} \\  
\midrule
\multicolumn{7}{c}{\textbf{CNN Encoder}} \\  
\midrule
Strides       & \multicolumn{6}{c}{5, 2, 2, 2, 2, 2, 2} \\  
Kernel Width  & \multicolumn{6}{c}{10, 3, 3, 3, 3, 2, 2} \\  
Channels      & \multicolumn{6}{c}{512} \\  
\midrule
\multicolumn{7}{c}{\textbf{Transformer}} \\  
\midrule
Depth ($l$)              & 24 & 24 & 24 & 4 & 4 & 4  \\  
Emb. Dim ($d_{\text{emb}}$) & 1024 & 1024 & 1024 & 1024 & 512 & 512 \\  
FFN Dim ($d_{\text{ffn}}$)  & 4096 & 4096 & 4096 & 2048 & 2048 & 2048 \\  
Attn. Heads ($h_{\text{attn}}$) & 16 & 16 & 16 & 16 & 16 & 16 \\  
\midrule
\multicolumn{7}{c}{\textbf{Projection}} \\  
\midrule
Dim. ($d_p$) & 768 & 768 & 768 & 768 & 768 & 768 \\  
\midrule
\multicolumn{7}{c}{\textbf{Params}} \\  
\midrule
\textit{in M} & 300 & 316 & 316 & 65 & 28 & 28 \\  
\bottomrule
\end{tabular}}
\end{table}

\textbf{Iteration 1 \& 2:} The HArnESS model is trained for three iterations using fairseq codebase \cite {ott2019fairseq}. The architecture for HArnESS-L, HArnESS-S and HArnESS-ST are shown in Table \ref{tab:arch_desc}. For the first two iterations, we used the architecture HArnESS-L, which consists of a 24-layers.
We train the HArnESS-L model on 23k
hours of Arabic-English audio on 24 H100 GPUs, with a batch size of at most 62.5 seconds of audio per GPU. The first iteration is trained for 500k steps, while the second iteration is trained for 700k steps. In the first iteration, we use 39-dimensional MFCC features and apply k-means clustering with 1000 clusters to generate labels. For the second iteration, to get the better targets, we extract latent representations from the 9th layer of the first iteration model and cluster them using k-means with 1000 clusters to generate new labels. \\
\textbf{Iteration 3:}
In the iteration $i=3$, we used the HArnESS-S and HArnESS-ST architectures, which have a shallower 4-layer transformer with reduced embedding dim 1024 and 512 respectively. We train HArnESS-S and HArnESS-ST for the third iteration on 1100 hours of arabic audio on 8 H100 GPUs, for 300k steps with a batch size of 75 seconds of audio per GPU. We extract features from the last transformer layer of the second iteration HArnESS-L for clustering (clusters=1000) and use those labels for training these two models. Hence, these two models can be seen as the third iteration models. 

\begin{figure}[!h]
    \centering    
    
    \includegraphics[scale=0.65]{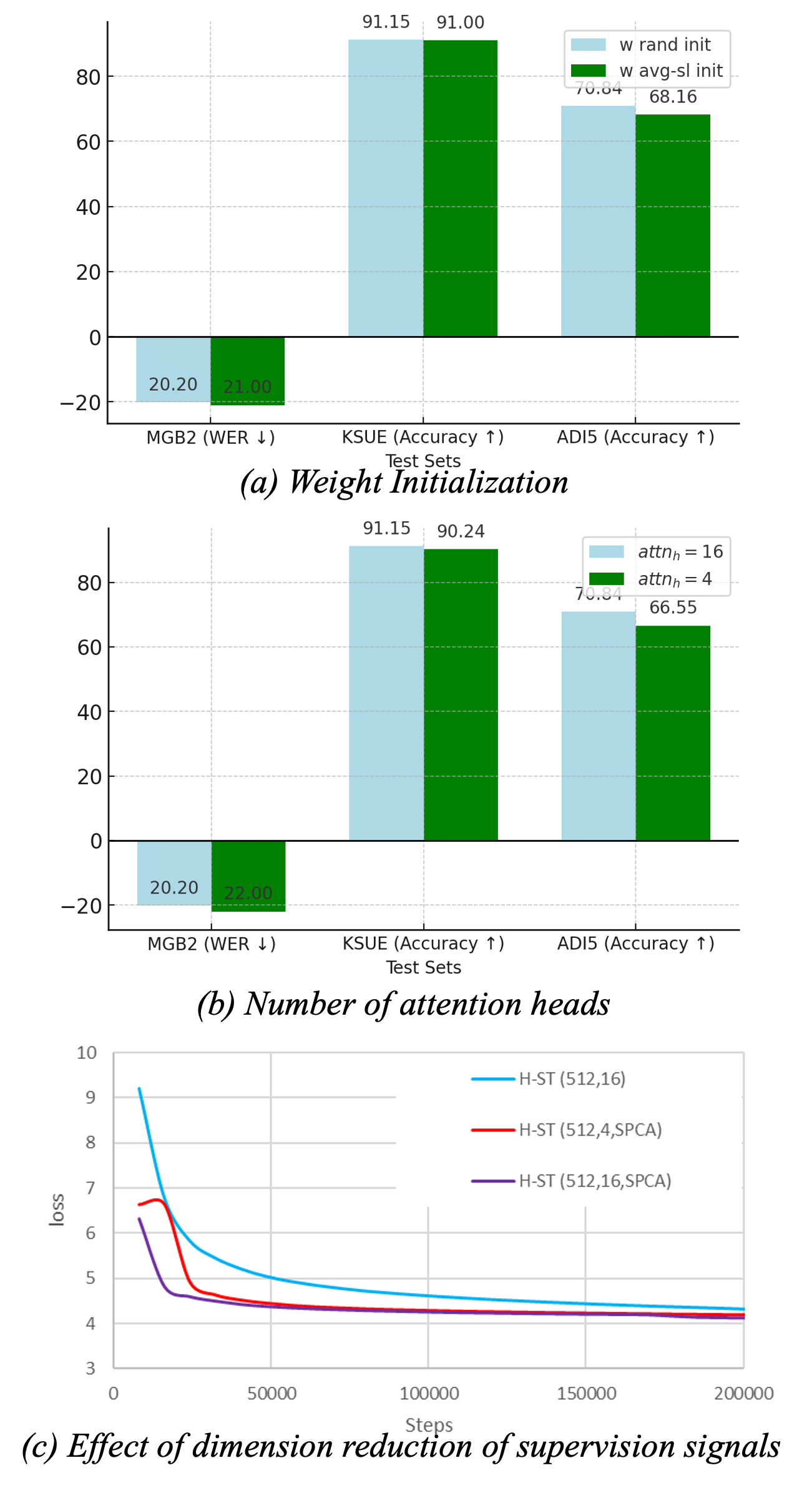}        \vspace{-0.3cm}
    \caption{(a) Performance of different downstream tasks based on the weight initialization strategy of the student; (b) performance for various $attn_h$; (c) Effects of PCA for supervision signal of teacher (SPCA) before clustering ($emb_d$, $attn_h$. }
    \label{fig:abl}
    \vspace{-0.15cm}
\end{figure}


\vspace{-10pt}
\subsection{Downstream Training}

For the downstream task, we utilized the SSL models as a feature extractor, and study its effectiveness for capturing rich representation. 
We averaged extracted embeddings from all the SSL layers and then passed them as a feature to the downstream network.


\subsubsection{DID and SER Architecture}
For classification tasks, we opt for a simple CNN with the self-attenion-based network. 
The network processes the input SSL features with three consecutive temporal CNN layers, with a kernel size of 5, and ReLU activations with a dropout of 0.4 for the generalization. Following a self-attention pooling is applied, and passed through a FF layer before passing it to the output layer. The hidden dimensions for all are set to 80.
Each model is trained with a batch size of 4 and ran for a total of 10K steps.



\subsubsection{Automatic Speech Recognition}
For ASR task, we trained an encoder-decoder architecture while optimizing joint CTC and attention loss.\footnote{Using ESPnet toolkit} The encoder comprises two layers of a conformer encoder followed by 2 layers of a transformer decoder, each layer incorporating 8 attention heads and 2048 linear units. The ASR model is trained for 70 epochs.


%


\section{Results}

 \paragraph*{Comparision with Upper Bound: SOTA Model} As an upper bound, we present the state-of-the-art (SOTA) performance for Arabic ASR, DID \cite{kulkarni-aldarmaki-2023-yet}, and SER models. The results in Table \ref{tab.rslt} show that both compressed HArnESS models and HArnESS-L outperform SOTA models in SER and DID, despite having a simpler architecture and shorter training time. This highlights the model's ability to capture rich speaker and paralinguistic information effectively.
For ASR, when compared to Fanar ASR \cite{fanarteam2025}, which is trained on 10K+ hours of dialectal data, HArnESS models, with only 300 hours of MSA fine-tuning, are just 5 (HArnESS-L) and 10 (HArnESS-S) points behind. 



\paragraph*{HArnESS-L {\em  vs.} Exiting SSLs for Arabic}
The HArnESS models outperform both HuBERT and XLS-R across all the Arabic tasks, demonstrating the importance of having language-specific models. Despite 
the similar scale of the model, HArnESS-L clearly benefits from the language-specific knowledge while outperforming a multilingual model itself. HArnESS-S and HArnESS-ST clearly outperforms the XLS-R indicating that the compressed HArnESS also captured abstract representation from the large model.


\paragraph*{Effects of different Structural Compression and design choices} 
For iteration $i=3$, we explored the impact of weight initialization in the student model and observed no significant change in downstream task performance (Figure \ref{fig:abl}), suggesting that initialization plays a minimal role at this stage of training.

With layer reduction, HArnESS-S achieves 79.4\% structural compression compared to HArnESS-L while maintaining strong performance across tasks. It outperforms multilingual and English models, demonstrating its efficiency in Arabic speech tasks despite compression. However, compared to HArnESS-L, we observe a WER increase of 4.7, a SER accuracy drop of 3.51, and a DID accuracy decline of 14.4, indicating that dialectal nuances become less distinguishable in shallower networks, making DID the most impacted task.

Further reducing attention heads from H-S ($attn=16$) to H-S$^*$ ($attn=4$)  results in an additional 26.15\% structural compression (from 65M to 48M parameters). While DID performance suffers the most, the effect on SER and WER remains minimal (Figure \ref{fig:abl}).

With embedding dimension reduction (Table \ref{tab.embred}), we observe a sharp drop in performance at $\Delta S=96.52\%$ compression relative to HArnESS-L, highlighting the limitations of excessive dimensionality reduction, which significantly degrades model performance.


\paragraph*{How Does Compressing the Supervision Signal Affect Performance?}
To examine the impact of compressing the supervision signal, we compare model loss during knowledge distillation in iteration $i=3$ with and without PCA applied to the teacher's supervision signal before clustering. Figure \ref{fig:abl} shows that PCA-based supervision converges faster than its counterparts, suggesting that reducing redundancy in the supervision signal improves training efficiency while maintaining effective knowledge transfer.


\begin{table}[h]
\centering
\caption{Performance Comparison for different embedding dimensions. $\Delta S$: Overall structural compression.}
\label{tab.embred}
\scalebox{0.85}{
\vspace{-0.3cm}
\begin{tabular}{l|ccc}
\toprule
\textbf{Test Sets} & \textbf{$emb_d$=1024} & \textbf{$emb_d$=512} & \textbf{$emb_d$=256} \\ 
\midrule
MGB2 (WER ↓)  & 20.2  & 23.20    & 22.3  \\  
KSUE (Acc ↑)  & 91.15\%  & 89.02\%  & 79.42\%  \\  
ADI5 (Acc ↑)  & 70.84\%  & 69.77\%  & 53.41\%  \\  
\midrule
\textbf{$\Delta S$} & 70.43\%  & 91.14\%  & 96.52\%  \\
\bottomrule

\end{tabular}}
\vspace{-0.3cm}
\end{table}




\section{Conclusion}
In this work, we introduced HArnESS, the first self-supervised Arabic-centric speech model family, designed to capture nuances of Arabic dialects. Through an iterative self-distillation approach, we transferred knowledge from large bilingual models to a shallow (and thin) student models while preserving Arabic-specific speech representations. Our experiments on Arabic ASR, SER, and DID tasks demonstrate that HArnESS achieves state-of-the-art or comparable results against existing multilingual models like HuBERT and XLS-R. The lightweight of HArnESS also makes it a highly efficient yet performance-compromised alternative. We will make available the lightweight models, and the benchmarking data for research purpose.



\bibliographystyle{IEEEtran}
\bibliography{mybib}

\end{document}